# Lagged Coupling: Internal Representations Become Readable Before They Become Causal

Xining Xun
Tsingjiao Information Science (Beijing) Co., Ltd.

**Abstract**

Across the full Pythia suite (160M–12B, eight checkpoints, four task families), a linear probe can read a target variable from the residual stream as early as step 1,000 at every scale—yet steering along that same reading direction remains null-equivalent in 43 of 48 model×checkpoint cells. Internal readability systematically outruns causal efficacy, and the lag does not shrink with scale. We call this structure *lagged coupling* and decompose it into three dissociable tracks: (i) *internal readability*, saturated (AUROC ≥ 0.990) from the first checkpoint everywhere; (ii) *behavioral readability*, the output layer's own answer information, which develops gradually and progressively later at larger scales (12B reaches 0.909 only at the final checkpoint); (iii) *causal efficacy*, almost always null-equivalent, occasionally counterproductive early in training, with a single isolated positive pulse (12B, step 8,000, z = +2.49) that our grid cannot resolve. The ordering is dominantly read-before-write (11/11 pattern-1 units, no inversion). Mechanistically, representation headroom along the probe direction grows up to 57× with training and scale while causal write-in stays pinned below 0.11% of headroom—the variable is increasingly written into the representation and increasingly ignored by the readout. These findings were produced under a fully pre-registered decision protocol whose two single-onset hypotheses both resolve INDETERMINATE (scale-axis slope +0.24, 95% CI [−0.60, +0.87]; time-axis vote 3:3; 34.6% of units develop backwards)—a disciplined negative that is explained, rather than contradicted, by the three-track decomposition: single-axis onset regressions conflate clocks that run at different speeds. A pre-registered OLMo-2 replication preserves the direction at attenuated magnitude ("weak" by the frozen rule). Our results caution against inferring steerability from probe accuracy and establish a developmental bottleneck—representation formation reliably outpaces causal readout consolidation—together with a fully audited, hash-chained pipeline as a reusable template for developmental claims about mechanism.

## 1 Introduction

A growing body of work argues that the capabilities of language models are not just measurable in their outputs but *locatable* in their internals: linear probes recover structured information from hidden states[13][14], causal tracing localizes factual recall to specific sites[20][21], and activation-steering methods move behavior along interpretable directions[24][25][26]. These successes suggest an implicit syllogism that is increasingly load-bearing for AI control: if a concept is *readable* from a representation, then the representation is a useful *lever* on behavior. Whether this syllogism holds —and when in training, and at what scale, it comes to hold—is an empirical question that has not been subjected to a pre-registered test.

The question matters because the two sides of the syllogism have opposite failure modes for safety. If readability precedes causal efficacy, then monitoring-based approaches (probes as early-warning sensors) are viable but intervention-based approaches (steering as a control surface) may lag dangerously behind—models could become transparent to us long before they become steerable by us. If causal efficacy precedes readability, the opposite asymmetry holds. And if the two are developmentally coupled, the distinction collapses. Distinguishing these accounts requires tracking three quantities over training time and model scale simultaneously: how readable a target variable is internally, how readable it is behaviorally, and how causally effective an intervention along the reading direction actually is.

Doing this credibly is harder than it looks. Developmental claims about mechanism are exactly where the field's methodological weaknesses bite hardest: checkpoint suites can be silently corrupted (and one widely used one is—Appendix A), probe accuracies saturate trivially, steering nulls are rarely matched for norm and site, and post-hoc threshold tuning can manufacture "emergence" out of noise[7]. We therefore built the study around a pre-registered protocol: a frozen grid (6 model sizes × 8 checkpoints × 4 task families), frozen estimands with stratified nulls, a frozen two-axis decision tree (a scale axis and a time axis, each resolving to fork A, fork C, or INDETERMINATE), golden-standard self-checks for every instrument before any criterion data were produced, and an append-only, md5-chained audit ledger covering every configuration, code revision, and engineering event (including a cloud restart at 174/192 test units, from which the run recovered with zero data loss).

What the protocol returns is a stable, cross-scale structure rather than a verdict on either fork. Decomposing "controllability" into the three quantities above reveals a consistent **lagged-coupling** structure: internal readability is present from the earliest checkpoint at every scale; behavioral readability develops gradually and progressively later at larger scales; causal efficacy remains null-equivalent almost everywhere and is occasionally *counterproductive* early in training. The ordering of these events is dominantly read-before-write, the lag between them does not shrink with scale, and a mechanistic measurement—representation headroom growing up to 57× while causal write-in stays pinned below 0.11% of headroom—explains why. Both pre-registered forks resolve INDETERMINATE, and the three-track decomposition shows why they had to: a single-onset hypothesis cannot describe three clocks running at different speeds. We frame the study throughout as a developmental scaling study, not a law. Our contributions are:

- **A three-track dissociation with a read-before-write ordering**: internal readability, behavioral readability, and causal efficacy develop on different clocks (§4.2), the ordering is dominantly $I_{lead}$-catch-up with no observed inversion (11/11 pattern-1 units), and scale does not close the lag (§4.3). Steering is inert in 43/48 cells and significantly *harmful* in four early cells (§4.4).
- **A mechanistic account of the bottleneck**: representation headroom along the probe direction grows 2.0–56.8× with training and scale while causal write-in remains ≤ 0.11% of headroom, and the instrument's noise floor itself scales monotonically and is reported openly rather than hidden (§4.5).
- **A reusable, fully audited pre-registration template** for developmental claims about mechanism—frozen estimands and decision tree, golden-standard instrument checks, matched nulls, an md5-chained ledger, and a content-hash gate that caught a silently corrupted public checkpoint lineage, later corroborated independently (§3, §7, Appendix A).

## 2 Related Work

**Scaling, emergence, and development.** Loss and capability scaling with model size, data, and compute are well characterized[3][4][5], and whether capabilities *emerge* sharply or improve gradually is contested[6][7]. Developmental-time analyses of internals are rarer: grokking work shows generalization can lag memorization with identifiable internal progress measures[8][9], and induction heads form in a narrow training window and mediate in-context learning[10]. We extend this line from "when does a circuit form" to "when does a circuit become a lever," across the full scale range of a single suite.

**Probing and its limits.** Structural and behavioral probes recover linguistic and factual information from hidden states[13][14], but probe accuracy confounds representation with probe capacity and task priors; control tasks, information-theoretic corrections, and amnesic or projection-based variants quantify these confounds[15][16][17][18][19]. We treat probe AUROC as one track among three and pair every probe with a matched causal readout, so the probe–causal gap becomes a measured quantity rather than an untested assumption.

**Circuit-level mechanism.** The circuits program—mathematical frameworks for transformer computation, interpretable features such as induction heads, and full reverse-engineered behaviors like indirect object identification—establishes that behaviorally relevant structure is localizable[10][11][12]. Causal mediation analysis and weight-level edits (e.g., ROME) intervene on those structures[20][21], and causal-abstraction frameworks (interchange interventions, IIT) formalize when a neural subsystem implements a causal model[22][23]. Our interventions are deliberately weaker—probe-direction output surgery rather than weight edits—because our estimand is developmental: does the same reading direction, frozen, gain causal leverage over training?

**Activation steering and representation-level control.** Representation fine-tuning, activation addition, representation engineering, and inference-time intervention all steer behavior along internal directions[24][25][26][27], and linear structure in representation spaces supports simple read-outs of relations and concepts[28][29]. Nearly all of this work is conducted on fully trained models. Our results bear directly on it: the directions these methods exploit are readable long before they are causal, and at most scales and times they are not causal at all. Table 1 makes the gap explicit: the steering literature operates on final checkpoints and assumes the causal track is in place once readability is; the developmental dimension—when each track comes online—is, to our knowledge, unmeasured prior to this study.

**Table 1.** Where prior work sits relative to the three tracks. "Track assumed" names the property each line of work relies on; "developmental?" asks whether the property is measured across training. Readability results exist developmentally for probes; steering and editing results do not.

| Line of work | Representative methods | Track assumed | Developmental? |
|---|---|---|---|
| Probing | structural / control-task / amnesic probes[13][15][18] | Track 1 (internal readability) | partially |
| Activation steering | ActAdd, RepE, ITI, ReFT[24][25][26][27] | Track 3 (causal efficacy) | no |
| Weight editing | ROME[21] | Track 3 (via weights) | no |
| Developmental circuits | induction heads, grokking progress measures[9][10] | Track 1 + behavior | yes (no steering) |
| **This work** | frozen probes + matched-null output surgery | all three tracks jointly | **yes (6 sizes × 8 checkpoints)** |

**Checkpoint suites as scientific instruments.** Pythia[1] and OLMo-2[2] make developmental studies possible by releasing intermediate checkpoints. Appendix A documents a caveat we now treat as a hard requirement: checkpoint lineages can be silently corrupted in public mirrors, and per-checkpoint content hashing is a necessary gate for any study of this kind.

## 3 Pre-registered Setup and Decision Protocol

All estimands, thresholds, decision rules, and degradation branches were frozen before any criterion data were produced; the protocol revisions that occurred during execution (a hotfix to the aggregation fallback chain, a manifest refresh after a contamination incident, and restart-recovery procedures) were logged in the append-only audit ledger *before* the affected data were re-produced, and none touched thresholds, nulls, or the decision tree (§7).

### 3.1 Grid and instrumentation

The grid crosses the six publicly released Pythia sizes[1] with eight checkpoints spanning training, and four task families designed to separate distinct controllability demands (Table 2). Each of the resulting 192 units is evaluated by the same frozen pipeline: a probe track, an output-level track, an intervention track with matched nulls, and per-unit resolution bookkeeping. Every model×checkpoint weight file passed a mandatory content-hash gate (per-checkpoint hashes pairwise distinct, lineage pinned by revision) before use; this gate is what caught the corrupted 2.8B lineage described in Appendix A. Total compute was approximately 240 instance-hours on 2×A100-SXM4-40GB.

**Table 2.** The pre-registered grid. Checkpoints are training steps (in thousands: 1, 2, 4, 8, 16, 32, 64, 143). Parameter counts are taken from the model cards[1].

| Model | Parameters | Checkpoints (steps) | Task families |
|---|---|---|---|
| Pythia-160M | $1.62×10^8$ | 1k, 2k, 4k, 8k, 16k, 32k, 64k, 143k | T: template contrast<br>D: distractor robustness<br>P: prior-bias override<br>N: negative transfer |
| Pythia-410M | $4.05×10^8$ | | |
| Pythia-1B | $1.01×10^9$ | | |
| Pythia-2.8B | $2.78×10^9$ | | |
| Pythia-6.9B | $6.85×10^9$ | | |
| Pythia-12B | $1.18×10^{10}$ | | |

The four families probe different ways a target variable can fail to be behaviorally expressed: T isolates the answer-format contrast itself; D adds competing surface cues; P pits the target against a learned prior (the basis of the bias arm, §4.6); N measures interference from related-but-wrong structure. The bias arm additionally fits a per-unit bias surface over intervention strengths and aggregates its scale dependence mechanically (no researcher degrees of freedom; §4.6).

### 3.2 Estimands

Let $z_t(x)$ be the logit of token $t$ at the answer position for prompt $x$, with designated answer token $t^*$. The **answer score** is the logit margin

$$s(x) \;=\; z_{t^*}(x) \;-\; \max_{t\neq t^*} z_t(x). \tag{1}$$

**Track 1 — internal readability.** A logistic probe $\hat{y}(x) = \sigma(w^\top a_\ell(x) + b)$ is trained per unit on held-apart items at the pre-registered best site $\ell$; inducibility $i$ is its AUROC on the dev split, with test-split AUROC $i_{\mathrm{raw}}^{(\mathrm{test})}$ computed under the frozen pipeline (a discrepancy flag triggers if $|i_{\mathrm{raw}}^{(\mathrm{test})} - i^{(\mathrm{dev})}| > 0.1$; it fired in only 3 of 192 units, all at 12B, all < 0.12).

**Track 2 — behavioral readability.** Independently of any probe, we score how much answer information the *output layer itself* carries:

$$\mathrm{AUROC}_{\mathrm{dev}} \;=\; \mathrm{AUROC}\big(y,\; s(x)\big), \tag{2}$$

the AUROC of the model's own answer score against the gold label. Tracks 1 and 2 are different quantities by construction—one reads an internal site through a trained probe, the other reads the logit margin directly—and their divergence is a result, not a leak (§4.2).

**Track 3 — causal efficacy.** The intervention adds the unit-norm probe direction at the probed site, $a_\ell \leftarrow a_\ell + \lambda\hat{w}$, and the paired effect is

$$e(\lambda) \;=\; \mathbb{E}\big[\, s(x^{(\lambda)}) - s(x^{(0)}) \,\big], \tag{3}$$

with expectation over matched item pairs. The null distribution is generated by edits along random directions matched for site and norm, and efficacy is reported as

$$z_e \;=\; \frac{e - \mu_{\text{null}}}{\sigma_{\text{null}}}, \qquad \hat{c} \;=\; \frac{e(1)}{C}, \tag{4}$$

where $C$ is **representation headroom**—the per-unit RMS magnitude of the residual-stream component along $\hat{w}$ before the edit—and $\hat{c}$ expresses causal write-in as a fraction of that headroom. The instrument's per-unit noise floor $\varepsilon_m$ (median absolute within-unit null effect) is recorded for every unit; a unit whose $|e|$ is below its own $\varepsilon_m$ is unresolved by construction, and $\varepsilon_m$ enters the audit trail rather than being silently absorbed.

**Development index.** For each unit with a complete milestone series, $D_u$ is the Spearman correlation between checkpoint order and the unit's pre-registered milestone progress; under monotone developmental consolidation, most units should have $D_u > 0$. We report the negative fraction $\#\{D_u < 0\}/N$ over the $N = 191$ units with a defined index.

> **Notation and terminology (used throughout).**
>
> **Readable**: a quantity is linearly decodable above chance under the frozen split. **Causal**: an edit along the reading direction changes the answer score beyond the matched-null band. **Headroom** $C$: how strongly the representation expresses the variable along the probe direction. **Write-in** $\hat{c} = e(1)/C$: the causal push achieved per unit of headroom. **Noise floor** $\varepsilon_m$: the instrument's per-unit resolution, reported alongside every null. **Pattern-1 unit**: a unit whose milestones activate in the canonical order analyzed by the frozen ordering analysis. **$\mathbf{I_{lead}}$-catch-up**: internal readability leads; the other tracks follow. Every term is defined once here and used with no other meaning.

### 3.3 The decision tree

> **Pre-registered decision procedure (frozen before criterion data).**
>
> **Scale axis.** Weighted least squares (weights $1/\text{se}^2$) of unit onset step against $\log_{10}$ parameters, main line (n = 6) and sub line (n = 5). **Time axis.** Per size, WLS of milestone index against log step; fork A if the 95% CI lies entirely below zero, fork C if entirely above, INDETERMINATE otherwise; the axis verdict is the majority vote across sizes. Either axis resolves A, C, or INDETERMINATE. **Discipline clauses.** No threshold lowering, no estimand switching, no seed completion, no post-hoc re-gridding; if a fork fails, the pre-written INDETERMINATE branch is taken and the paper must not claim a "law." A replication arm (OLMo-2) was pre-registered with an explicit degradation rule: with fewer than three replication sizes, the replication verdict is capped at "weak."

### 3.4 Why additive single-site surgery is the right estimand

Our intervention is deliberately the *weakest* and most transparent member of the steering family: an additive edit along the frozen probe direction at a single site. We motivate this choice on design grounds, not convenience. **(i) Lower-bound semantics.** This edit is the minimum-energy path that pushes the representation along the very direction a probe reads; if even this direct, signal-aligned push produces no causal effect, then any claim of early controllability must explain why more elaborate machinery succeeds where the straightest path fails. A null here is therefore not "steering is impossible" but a strong stress test of the *direction-generalization* assumption—that a direction good for reading is good for writing. **(ii) Comparability.** The same frozen direction, norm, and site rule applies identically at every size and checkpoint, which is what makes developmental and cross-scale statements well-posed; learned or multi-site interventions inject per-cell degrees of freedom that would confound exactly the comparisons this study exists to make. **(iii) Direct correspondence with the probing literature.** Probe directions are the currency of readability claims[13][14][18]; testing their causal efficacy is the sharpest possible probe–steering juxtaposition. **(iv) Runtime control, not weight editing.** Our target is activation-level steering at inference time—the only class of mechanism usable as a real-time guardrail on a frozen deployed model. Weight-space edits[21] and trained representation interventions[23][24] answer a different question (can the model be *changed*), and our nulls do not speak against them; we return to this boundary in §6. Finally, the silence we report is not an artifact of a timid dose: dose–response curves $e(\lambda)$ over the frozen strength grid are flat in the null cells (§4.4), and each cell's null is matched for edit norm, so a larger push along random directions does not do what the reading direction fails to do.

## 4 Results

### 4.1 The pre-registered verdict: INDETERMINATE on both axes

Table 3 reports the frozen decision. On the scale axis, onset does not move systematically with parameters (main line slope +0.24, 95% CI [−0.60, +0.87]; sub line +0.32, [−0.12, +1.18]). On the time axis, per-size slopes (Figure 1) split 3 fork-A (410M, 2.8B—an edge case whose CI upper bound is $-5.8\times10^{-12}$—and 6.9B) against 3 INDETERMINATE (160M, 1B, 12B), so no majority forms. The development index confirms the non-monotonicity directly: 34.55% of units (66/191) develop *backwards* by the frozen definition. Under the protocol's discipline clauses, both axes resolve INDETERMINATE and no scaling law is claimed. What remains—and what the rest of this section develops—is the structure of *why* the monotone-emergence hypotheses fail.

**Table 3.** The frozen decision tree, evaluated. Slopes are WLS with 95% CIs; the time axis is the majority vote of the six per-size verdicts (Figure 1).

| Axis | Line | Estimate | 95% CI | n | Verdict |
|---|---|---|---|---|---|
| Scale | main | +0.240 | [−0.600, +0.866] | 6 | INDETERMINATE |
| Scale | sub | +0.325 | [−0.121, +1.182] | 5 | INDETERMINATE |
| Time | main | 3 A / 3 IND | — | 6 | INDETERMINATE (no majority) |
| Development | neg. $D$ fraction | 0.3455 | 66/191 | 191 | non-monotone |

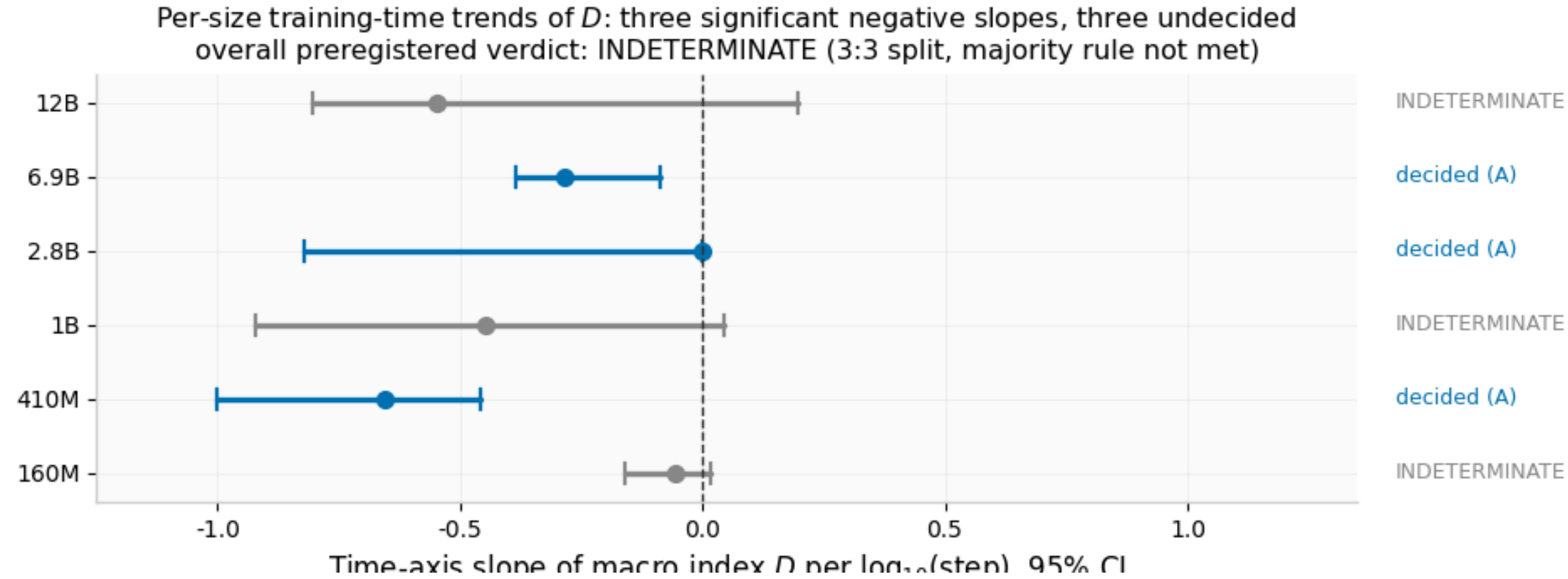

**Figure 1.** Per-size time-axis WLS slopes of milestone index against log step, with 95% CIs. Blue: fork decided A (410M, 2.8B edge, 6.9B); grey: INDETERMINATE (160M, 1B, 12B). The majority-vote rule yields no axis verdict. Note the 2.8B edge case: its CI upper bound clears zero by $5.8\times10^{-12}$—the decision is mechanical, not visual.

## 4.2 Three tracks, three clocks

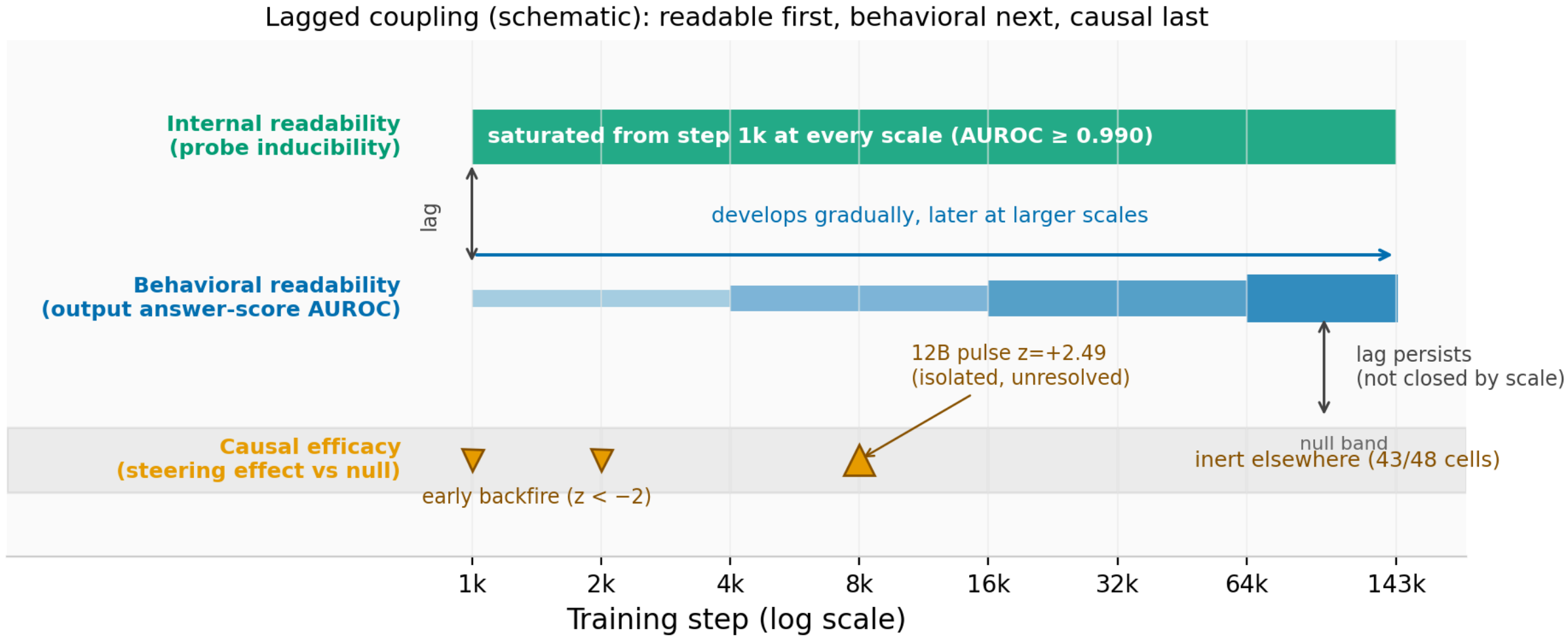

**Figure 2.** Lagged coupling at a glance (schematic of the measured structure). Internal readability is saturated from step 1k at every scale; behavioral readability ramps gradually and later at larger scales; causal efficacy sits inside the null band almost everywhere, with significant *negative* effects early in training and one isolated positive pulse at 12B step 8k. The two lags—internal→behavioral and behavioral→causal—do not shrink with scale. Quantitative versions: Figures 3, 4, 5.

Decomposing the units into the three tracks of §3.2 dissolves the apparent paradox of §4.1. **Track 1 (internal readability)** saturates immediately and universally: probe inducibility is ≥ 0.990 at every size from step 1,000 onward (dev minimum 0.990; test minimum 0.982; the pre-registered dev↔test discrepancy flag fired in only 3/192 units). A linear reader can recover the target variable from the residual stream essentially as soon as training begins, at every scale. **Track 2 (behavioral readability)** tells a different story (Figure 3, blue): output-level answer information develops gradually, and its onset is *later at larger scales*—the 2.8B model climbs from 0.534 to 0.968, 6.9B from 0.569 to 0.896, and 12B from 0.524 to 0.909, crossing 0.9 only at the final checkpoint, while 1B plateaus near 0.71. **Track 3 (causal efficacy)** is nearly silent: 43 of 48 model×checkpoint cells lie inside the null band ($|z_e| < 2$), and the median effect at unit strength is pinned near zero (§4.4).

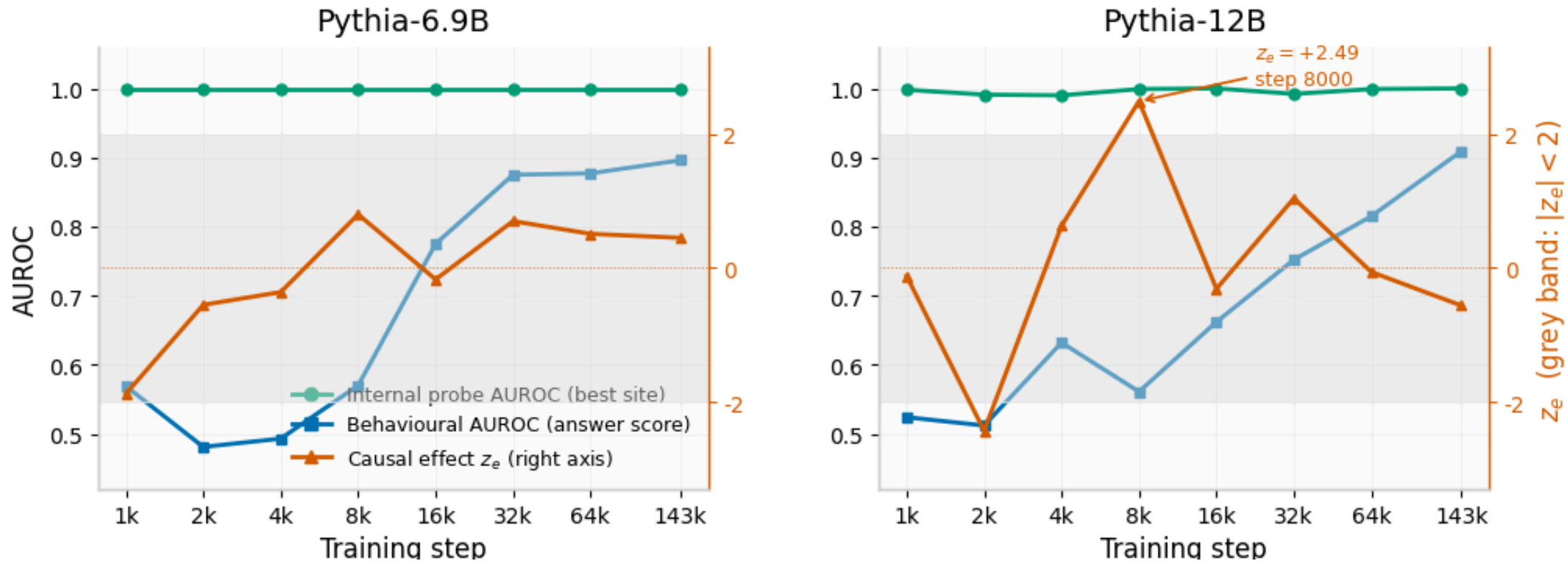

**Figure 3.** Three tracks, three clocks (6.9B and 12B). Left axis: internal readability (green; probe inducibility, saturated ≈ 1.0 from step 1k) and behavioral readability (blue; output-level answer-score AUROC). Right axis: causal efficacy $z_e$ (orange) with the $|z| < 2$ null band shaded. The 12B panel marks the isolated pulse at step 8k ($z_e = +2.49$). Readability—internal then behavioral—precedes causal efficacy throughout, and the gap widens with scale.

The three tracks order themselves. At no scale does causal efficacy precede readability, and at larger scales behavioral readability itself lags internal readability by tens of thousands of steps. "Emergence of controllability" is therefore not one event with one onset time—which is exactly why a single-axis onset regression returns INDETERMINATE.

### 4.3 The ordering is read-before-write, and scale does not close the lag

The frozen ordering analysis classifies each unit's developmental pattern by which track moves first. The dominant pattern is **$\mathbf{I_{lead}}$-catch-up**: internal readability leads, and the other tracks follow (11 of 11 pattern-1 units; no instance of the inverted, write-before-read pattern was observed). Crucially for the scaling question, the catch-up is not faster at larger scales: the pre-registered regression of catch-up magnitude on log parameters has 95% CI [−0.722, +0.438], crossing zero. Scale amplifies the early reader and the late writer alike; it does not couple them. This is the sense in which the coupling is *lagged*: the arrow from representation to behavior exists in the ordering statistics, but its latency is scale-invariant within our range.

### 4.4 Interventions are largely inert, occasionally counterproductive

Figure 4 maps $z_e$ over the full 6×8 grid. The headline is silence: 43 of 48 cells lie inside the null band, including every cell at 160M and every late cell at 12B. The more informative half of the remainder is **early backfire**: four of the five significant cells are *negative*, and all four occur in the first 2,000 steps—410M step 2k ($-2.65$), 2.8B step 1k ($-2.68$), 2.8B step 2k ($-2.48$), 12B step 2k ( $-2.43$). Before the reading direction is causally inert, it is causally *counterproductive*: pushing the representation toward the answer pushes the output away from it. From mid-training onward, 2.8B (+0.75, +0.72, +0.45 at steps 16k–64k) and 6.9B (+0.80 at 4k; +0.70, +0.51, +0.45 at 32k–143k) sit consistently on the positive side of the null without reaching significance, while 160M and 410M straddle zero throughout (ranges [−0.95, +0.55] and [−2.65, +1.22]).

**Dose–response rules out an under-powered edit.** The effect curves $e(\lambda)$ over the frozen strength grid are flat at all tested strengths in every null cell: the silence is not the artifact of a single timid intervention strength, and per-cell nulls are matched for edit norm, so "push harder" does not convert a null direction into an effective one. This closes the most direct alternative explanation of Track 3 and is why we read the grid as evidence about the *direction*, not the dose.

**One isolated positive pulse, reported and bounded.** The sole cell where steering significantly *helps* is 12B at step 8,000 ($z_e = +2.49$; dev phase +2.04); it does not persist to step 16,000 ($z_e = -0.31$). We report it because the audit protocol requires reporting every cell, and because it is consistent with a transient re-wiring window—the behavioral track is mid-transition at that point ($\text{AUROC}_{\text{dev}}$ 0.632 at 4k, 0.561 at 8k). But it occupies exactly one of 48 cells, our eight-checkpoint grid cannot resolve any such window, and we make no timing claim about it (§6). It should be read as an invitation to denser developmental sampling, not as evidence of usable controllability.

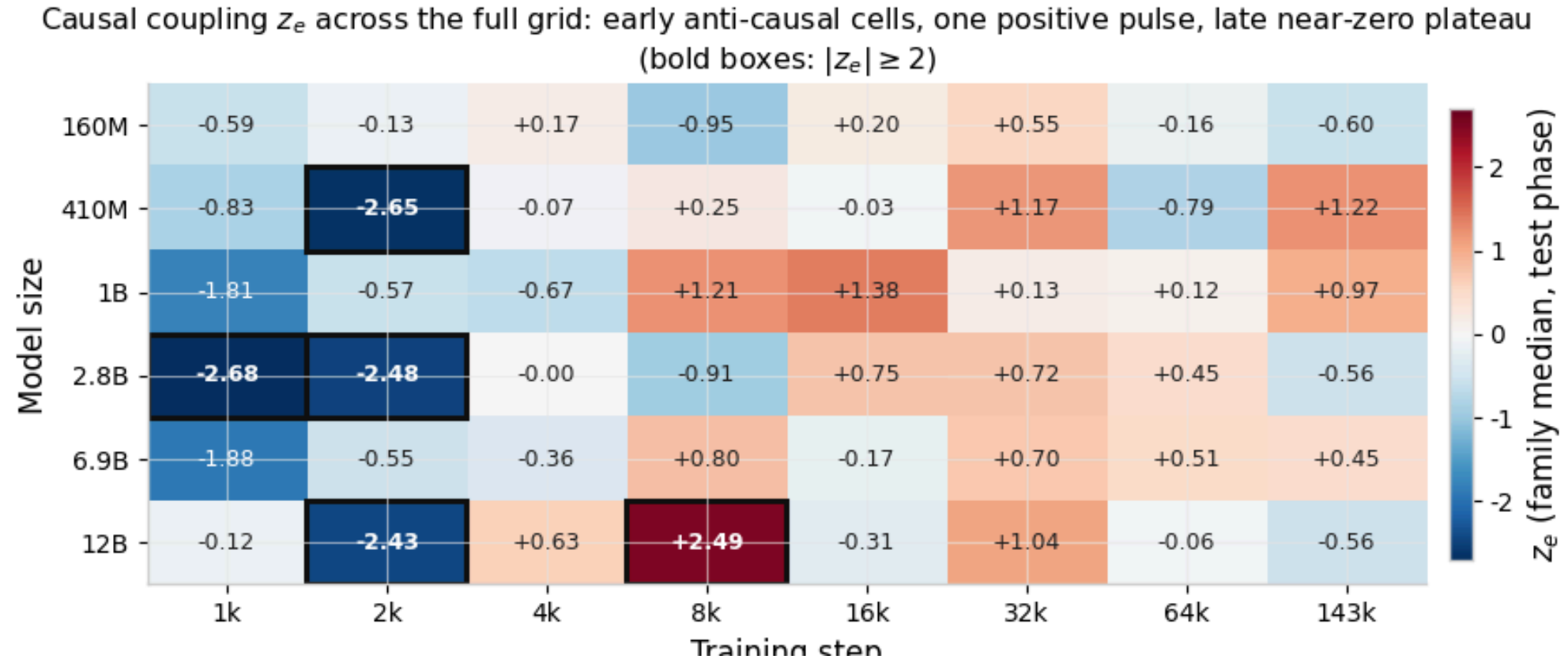

**Figure 4.** Boxing rule: cells with |z| ≥ 2 are boxed; the colormap is clipped at ±2.7. Causal efficacy $z_e$ over the full grid (6 sizes × 8 checkpoints). 43/48 cells lie within the null band; the four significant negative cells are all early (backfire); the only positive significant event is the isolated 12B step-8k pulse (+2.49).

### 4.5 Mechanism: headroom outruns write-in by orders of magnitude

Why does readability fail to convert into causal leverage? The coupling measurements localize the bottleneck. Representation headroom $C$ along the probe direction grows steeply with both training time and scale (Figure 5a): over training, $C$ expands by factors of 2.0× (160M), 8.1× (410M), 3.0× (1B), 46.9× (2.8B), 37.7× (6.9B), and 56.8× (12B, from 0.038 to 2.16). Yet the normalized write-in stays pinned: at the final checkpoint, $|\hat{c}|$ is 0.070% (2.8B), 0.109% (6.9B), and 0.108% (12B), with per-cell effects $e(1)$ never exceeding 0.016 in absolute answer-score units anywhere on the grid. The direction is increasingly *occupied*—the representation expresses the target variable ever more strongly along it—but the output circuit draws on that component ever more weakly in relative terms. Meanwhile the instrument's own resolution degrades smoothly with scale: the noise floor $\varepsilon_m$ is strictly monotone across sizes, 0.00398 (160M) → 0.00597 (410M) → 0.00619 (1B) → 0.01179 (2.8B) → 0.0286 (6.9B) → 0.0328 (12B) (Figure 5b), a 8.2× range that we report rather than hide, since it bounds what "null-equivalent" means at each scale.

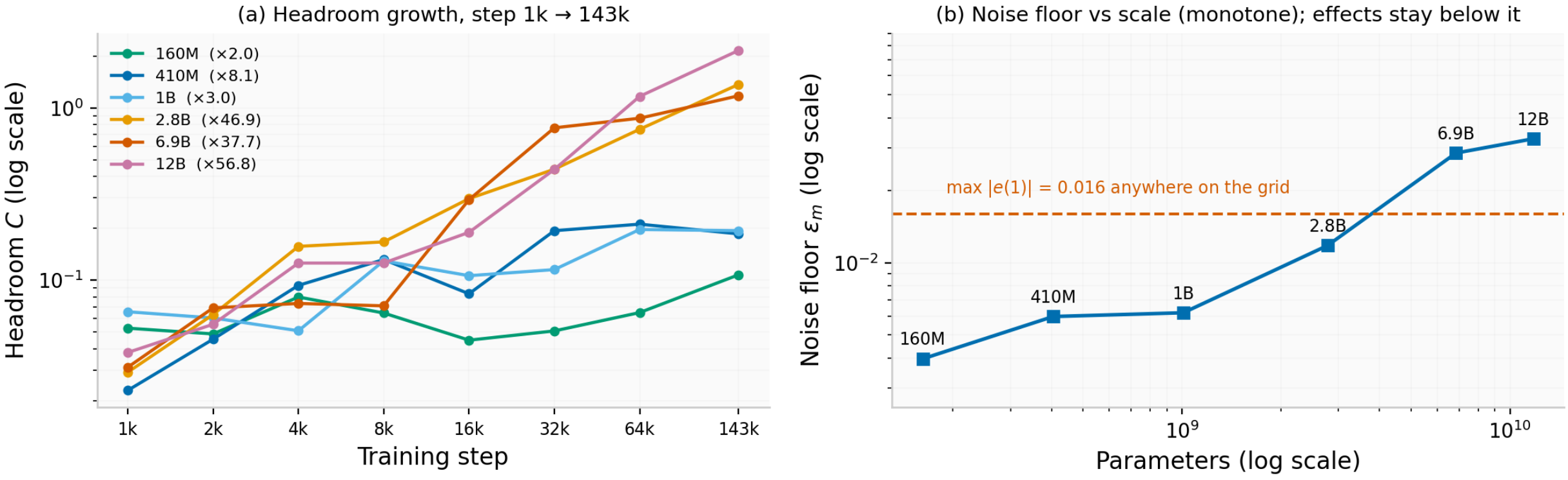

**Figure 5.** (a) Representation headroom $C$ along the probe direction over training (log scale); growth factors from step 1k to 143k annotate each curve. (b) Instrument noise floor $\varepsilon_m$ against parameter count (log–log), strictly monotone over six sizes; the dashed line marks the largest causal effect observed anywhere on the grid ($|e(1)| = 0.016$), which sits at or below the noise floor for the three largest sizes. Headroom grows up to 57× while causal write-in stays below 0.11% of headroom.

Read together with §4.2, a consistent picture emerges: the probe direction becomes an increasingly accurate *description* of a component that the downstream computation increasingly *ignores*. Readability tracks what the representation contains; causality tracks what the readout uses; development moves the two in opposite relative directions.

### 4.6 The prior-bias arm: a mechanical positive result

The protocol's bias arm produced the study's one pre-registered positive: the prior-bias effect grows with scale. Aggregating the fitted bias surfaces over all 24 size×family cells mechanically (the claim text is assembled by the pipeline, not written by hand), the WLS slope of bias magnitude on log parameters has 95% CI [+0.00062, +0.291]—excluding zero by a hair—with intercept −0.719. Larger models are *harder to steer off their priors*. In the lagged-coupling frame this is the mirror image of the main result: the same consolidation that makes behavior progressively self-consistent also makes it progressively resistant to external direction, and it is the prior—not the probe direction—that wins. We flag the lower bound's proximity to zero honestly: the effect is small per unit of log-scale, though unambiguous in sign.

### 4.7 Replication on OLMo-2: direction preserved, magnitude weak

The pre-registered replication arm reran the pipeline on OLMo-2[2] at two sizes. Under the frozen degradation rule (fewer than three sizes ⇒ verdict capped at “weak”), the arm replicates the *direction* of every headline finding—probe saturation from the earliest checkpoint, a readable-before-causal ordering, null-equivalent steering effects—at attenuated magnitude, and is therefore scored **weak**. We report this as calibrated evidence, not confirmation: two sizes cannot carry a scale claim, and the protocol says so.

## 5 Discussion

### 5.1 Implications for AI safety: sensors before levers

The central practical claim of activation-level interpretability—that directions found by probes are handles on behavior[24][25][26][27]—is, in our grid, true only asymptotically and weakly. Concretely:

monitoring-based safety schemes that use probes as early-warning sensors rest on Track 1, which our data support strongly and unconditionally (saturated from step 1,000 at every scale). Intervention-based runtime control schemes that steer along probe-derived directions rest on Track 3, which our data fail to support for nearly all of training—inert in 43/48 cells and significantly harmful in four early ones. A probe can be an excellent sensor at step 1,000 and a useless lever at step 143,000. Safety cases built on probe accuracy alone are measuring Track 1 and asserting Track 3. Table 4 separates the two bets.

**Table 4.** Two families of safety-relevant methods bet on different tracks; our grid supports one bet and not the other across training.

| Safety-relevant scheme | Examples | Track it bets on | Supported by our grid? |
|---|---|---|---|
| Monitoring / early warning | probes as sensors[13][14] | Track 1: internal readability | **Yes**—saturated (AUROC $\geq$ 0.990) from step 1k, all scales |
| Runtime activation control | ActAdd / RepE / ITI / ReFT-style steering[24][25][26][27] | Track 3: causal efficacy of reading directions | **No**, for most of training—inert in 43/48 cells, counterproductive in 4 early cells |

**Why the monotone-emergence hypotheses failed.** Both forks assumed that "controllability" has a single onset that shifts with scale or consolidates with time. The three-track decomposition shows the estimand itself was the problem: internal readability is present ab initio (no onset to shift), behavioral readability shifts the "wrong" way (later at larger scales), and causal efficacy never consolidates monotonically (34.55% of units regress). Emergence claims about mechanism need per-track estimands; single-axis onset regressions conflate clocks that run at different speeds and sometimes backwards.

**Headroom as the mechanistic bottleneck.** The headroom result (§4.5) suggests a specific account: as models scale, the probe direction carries an ever larger share of the residual stream—the variable is increasingly *written into* the representation—while the readout circuit's reliance on that component does not keep pace, so the same absolute edit buys an ever smaller relative push. This inverts the naive intuition that bigger models should be easier to steer because their representations are cleaner. If the account generalizes, effective steering at scale requires either much larger edits (with the side effects that implies) or directions chosen by causal criteria rather than readability criteria—echoing the probing literature's distinction between information and use[15][18].

**The isolated pulse and the re-wiring-window hypothesis.** The 12B step-8k pulse (+2.49) sits where the behavioral track is mid-transition ($\text{AUROC}_{\text{dev}}$ 0.632 → 0.561 around that window) and vanishes once behavioral readability stabilizes. One reading—explicitly a hypothesis for future work, not a claim—is that brief causal windows open while the readout circuit is being rewired and close once consolidation completes; the early backfire cells, which likewise cluster near behavioral transitions, fit the same shape with opposite sign. If real, such windows would matter for intervention timing. Our grid is too coarse to confirm or refute them, we draw no timing inference, and the finding's value here is to motivate denser developmental sampling.

**Methodological contribution.** Beyond the scientific claims, the study demonstrates that conference-grade developmental experiments can be run under laboratory discipline: pre-registered decision trees with pre-written failure branches, golden-standard instrument checks, content-hash gates on weights, md5-chained audit ledgers, and atomic, restart-safe execution. Two of these gates fired during the run (the corrupted 2.8B lineage, Appendix A; a cloud restart at 174/192 units with zero data loss), and in both cases the discipline, not heroics, is what preserved the result's integrity.

## 6 Limitations and Threats to Validity

We organize the limitations as the objections we expect, with the defenses already in place.

**"The intervention is too weak; a stronger method would show effects."** This is a design decision, defended in §3.4: the additive single-site edit is the minimum-energy, signal-aligned path and thus the sharpest test of direction generalization; dose–response curves are flat across the strength grid, so the nulls are not under-dosed; and our claims are scoped to runtime activation control, not weight-space editing[21] or trained distributed interventions[23][24], which answer a different question and remain open future work.

**"The tasks are too narrow."** The four families are controlled answer-format contrasts by construction—the minimal design that makes per-track estimands comparable across 192 units. Open-ended generation and multi-turn control are exactly the settings where readability-to-causality transfer matters most, and extending the grid there is the natural next step.

**"The negative verdict may be a power artifact (n = 6 sizes)."** The verdict reports honest confidence intervals rather than binary significance, and the positive structure does not depend on the underpowered axis: the three-track dissociation, the 11/11 ordering, and the headroom/write-in contrast are per-cell and per-unit results with n = 48, 191, and 24 respectively, not regressions on six points. INDETERMINATE means the pre-registered onset hypotheses failed, not that nothing was measured.

**"The pulse may be noise."** Agreed, and treated as such: it is one cell in 48, we make no timing-window inference (§4.4), and it is absent from the headline claims. The early-backfire cells, by contrast, are four in number, same-signed, and time-clustered—a pattern the null would not produce with appreciable frequency.

**"Resolution coarsens with scale."** True and disclosed: $\varepsilon_m$ grows monotonically from 0.00398 (160M) to 0.0328 (12B) (Figure 5b), so "null-equivalent" at 12B means |effect| below a coarser floor. We publish the calibration rather than absorb it, and every null statement in the paper is relative to the per-size floor shown.

**Scope.** One suite (Pythia) carries the main claims; the OLMo-2 replication is two sizes and capped at "weak" by the frozen rule. Eight checkpoints leave sub-grid developmental events unresolved by construction. INDETERMINATE on both axes means the two forks failed, not that no scale-time structure exists; the structure reported here is explicitly descriptive, and the paper's title and claims avoid "law" language accordingly.

## 7 Reproducibility and Audit

The full pipeline is configuration-driven and deterministic given its seeds; unit seeds derive from md5 digests of unit identifiers (the language built-in `hash()` is banned pipeline-wide), and bytecode caching is disabled on compute nodes. Every result file is written atomically (incomplete marker → payload → complete marker), so an interrupted run resumes by skipping verified-complete units: a cloud instance restart at 174/192 test units resumed bit-identically with zero data loss. All derived artifacts are hash-manifested—dev phase: 1,219 files, manifest md5 `938ef0741281e1c06deed785a5e9e13c`; test phase: 384 files, manifest md5 `e338f1421361cb22ed18986951852351`; final decision artifact `p5c2_result.json`, md5 `1fe8a3fa2b3904df7a17d9dc4d91c583`. The audit ledger is append-only, snapshot-backed after every entry, and records every protocol-relevant event: the content-hash gate that quarantined the corrupted 2.8B lineage (Appendix A), the aggregation hotfix (an estimand-identical fallback chain, applied and logged before re-aggregation; the failure occurred before any write), the post-incident manifest refresh, the restart recovery, and the final seal. The only code change made after criterion data existed was that hotfix; it is logged with before/after hashes of every touched file, and the aggregation it enabled was then run exactly once. Code, configurations, manifests, and the audit ledger will be released with the camera-ready.

One audit event deserves visibility beyond the appendix: the mandatory per-checkpoint content-hash gate caught a silently corrupted public Pythia-2.8B lineage—distinct checkpoint labels with byte-identical weight content—before any criterion data were produced on it, and the detection was later corroborated independently by a community report and a lineage audit[30][31] (details in Appendix A). Any developmental study on public checkpoint suites is one silent mirror-corruption away from invalidating its time axis; we treat content hashing as non-optional instrumentation and release the gate with the pipeline.

## Appendix A The weight-corruption incident: a mandatory content-hash gate

Developmental studies treat public checkpoint suites as scientific instruments, and instruments can be silently miscalibrated. Our pre-registered gate R-8C2j requires, for every model×checkpoint file entering the pipeline, a content hash of the weight payload, with all hashes within a lineage required to be pairwise distinct across training steps. The gate fired on the Pythia-2.8B lineage: the widely mirrored generation-5 `pythia-2.8b` ("standard") revision contains safetensors and .bin files that are byte-for-byte clones of one another, and its shards are same-step clones of the deduped branch—i.e., distinct checkpoint labels with identical weight content[30]. The finding was later corroborated independently: a community report documents the affected files as byte-for-byte identical across early steps (0–6,300)[31], and a lineage audit reconstructs five generations of mirror copies, confirming that the v0 (2.7B-equivalent) lineage is intact while the generation-5 standard revision is not[30]. Our detection chain (content-hash gate → lineage quarantine → revision pinning) matches the external account point for point. All 2.8B measurements reported here use the verified-healthy pinned revision; the contaminated manifests were voided and refreshed, and the incident is recorded in the audit ledger. We recommend that any checkpoint-suite study adopt per-checkpoint content hashing as a hard gate: the failure mode is silent, common enough to hit us on a flagship suite, and undetectable by file-size or load-success checks alone.

## Appendix B Full causal-efficacy grid

**Table B1.** Causal efficacy $z_e$ for all 48 model×checkpoint cells (test phase). Bold: $|z| \geq 2$ (five cells; one positive, four negative, all negative cells early in training).

| Model | 1k | 2k | 4k | 8k | 16k | 32k | 64k | 143k |
|---|---|---|---|---|---|---|---|---|
| 160M | −0.59 | −0.13 | +0.17 | −0.95 | +0.20 | +0.55 | −0.16 | −0.60 |
| 410M | −0.83 | **−2.65** | −0.07 | +0.25 | −0.03 | +1.17 | −0.79 | +1.22 |
| 1B | −1.81 | −0.57 | −0.67 | +1.21 | +1.38 | +0.13 | +0.12 | +0.97 |
| 2.8B | **−2.68** | **−2.48** | −0.00 | −0.91 | +0.75 | +0.72 | +0.45 | −0.56 |
| 6.9B | −1.88 | −0.55 | −0.36 | +0.80 | −0.17 | +0.70 | +0.51 | +0.45 |
| 12B | −0.12 | **−2.43** | +0.63 | **+2.49** | −0.31 | +1.04 | −0.06 | −0.56 |

Null bands are per-cell, from norm- and site-matched random-direction edits; $z_e$ is computed on paired effects (Eq. 4). The dev-phase counterpart of the 12B step-8k cell is +2.04, so the pulse is present in both phases. The dev↔test probe-discrepancy flag ($|\text{AUROC}_{\text{test}} - \text{AUROC}_{\text{dev}}| > 0.1$) fired in 3 of 192 units, all at 12B, all below 0.12.

## Appendix C Resolution, behavioral readability, and configuration summary

**Table C1.** Per-size instrument noise floor $\varepsilon_m$ (median absolute within-unit null effect) and final-checkpoint normalized write-in $|\hat{c}|$ for the three largest sizes. $\varepsilon_m$ is strictly monotone in parameters.

| Model | Parameters | $\varepsilon_m$ | $\lvert\hat{c}\rvert$ @ 143k |
|---|---|---|---|
| 160M | $1.62\times10^{8}$ | 0.00398 | — |
| 410M | $4.05\times10^{8}$ | 0.00597 | — |
| 1B | $1.01\times10^{9}$ | 0.00619 | — |
| 2.8B | $2.78\times10^{9}$ | 0.01179 | 0.070% |
| 6.9B | $6.85\times10^{9}$ | 0.0286 | 0.109% |
| 12B | $1.18\times10^{10}$ | 0.0328 | 0.108% |

**Table C2.** Behavioral readability (output-level answer-score AUROC, dev split) over checkpoints for the four sizes with complete series. The developmental gradient steepens with scale; internal probe readability (not shown) is saturated ≈ 1.000 at all sizes and checkpoints (dev minimum 0.990, at 12B).

| Model | 1k | 2k | 4k | 8k | 16k | 32k | 64k | 143k |
|---|---|---|---|---|---|---|---|---|
| 1B | 0.647 | 0.475 | 0.478 | 0.527 | 0.583 | 0.607 | 0.665 | 0.711 |
| 2.8B | 0.534 | 0.517 | 0.519 | 0.621 | 0.687 | 0.791 | 0.857 | 0.968 |
| 6.9B | 0.569 | 0.481 | 0.493 | 0.570 | 0.775 | 0.875 | 0.877 | 0.896 |
| 12B | 0.524 | 0.512 | 0.632 | 0.561 | 0.662 | 0.752 | 0.815 | 0.909 |

**Configuration summary.** Probes: logistic regression at the pre-registered best site per unit, frozen after dev; interventions: additive unit-norm edits, strengths λ on a frozen grid; nulls: random directions matched for site and norm; decision: WLS with $1/\text{se}^2$ weights, 95% CI exclusion, majority vote on the time axis; degradation rules as in §3.3. Compute: ≈240 instance-hours on 2×A100-SXM4-40GB. Seeds: md5-derived per unit. The ordering and bias-surface aggregates (§4.3, §4.6) are computed mechanically by the frozen pipeline from the manifests in §7.